\documentclass[runningheads]{llncs}
\usepackage{graphicx}

\usepackage{tikz}
\usepackage{comment}
\usepackage{amsmath,amssymb} 
\usepackage{color}
\usepackage{kotex}
\usepackage[accsupp]{axessibility}  

\begin{document}
\pagestyle{headings}
\mainmatter
\def\ECCVSubNumber{100}  

\title{2nd Place Solution to ECCV 2022 Challenge: \\ Transformer-based Action recognition in hand-object interacting scenarios} 

\titlerunning{Transformer-based Action recognition} 

%
\author{Hoseong Cho and Seungryul Baek}

\authorrunning{Cho et al.}
%
\institute{Ulsan National Institute of Science and Technology (UNIST), South Korea\\
\email{\{hoseong, srbaek\}@unist.ac.kr}}


\maketitle

\begin{abstract}
This report describes the 2nd place solution to the ECCV 2022 Human Body, Hands, and Activities (HBHA) from Egocentric and Multi-view Cameras Challenge: Action Recognition. This challenge aims to recognize hand-object interaction in an egocentric view. We propose a framework that estimates keypoints of two hands and an object with a Transformer-based keypoint estimator and recognizes actions based on the estimated keypoints. We achieved a top-1 accuracy of 87.19\% on the testset.
\end{abstract}

\section{Introduction}
\label{sec:intro}
In augmented reality (AR), virtual reality (VR) and human-computer interaction, egocentric perception of humans is a crucial component. Since previous works have primarily focused on single hands~\cite{baek2018augmented,baek2019pushing,kim2021end} and object interaction scenarios~\cite{baek2020weakly,park2022handoccnet}, most of the datasets~\cite{chao2021dexycb,garcia2018first,hampali2020honnotate} included only one-hand and an object interaction. In addition, while recent progress has been made in video comprehension and action recognition, most datasets are focusing on actions captured in the third viewpoint. The H2O dataset~\cite{kwon2021h2o} provides interaction of two hands and an object in multi-views including the egocentric viewpoint. This challenge aims to recognize the interaction of hand and object. In the action recognition, the dynamics of hand and object keypoints contain considerable information~\cite{duan2022revisiting,yan2018spatial}. Therefore, we propose a framework that predicts the keypoints of two hands and an object for each single frame and uses the result as an clue to the temporal module. Recently in the field of computer vision, the Transformer shows significant performance in various tasks. We adopt Transformer-based architecture for both keypoints estimator and action classifier and the proposed architecture achieved the top-1 accuracy of 87.19$\%$ on H2O dataset.

\begin{figure*}[!ht]
\centering
\includegraphics[width=1\linewidth]{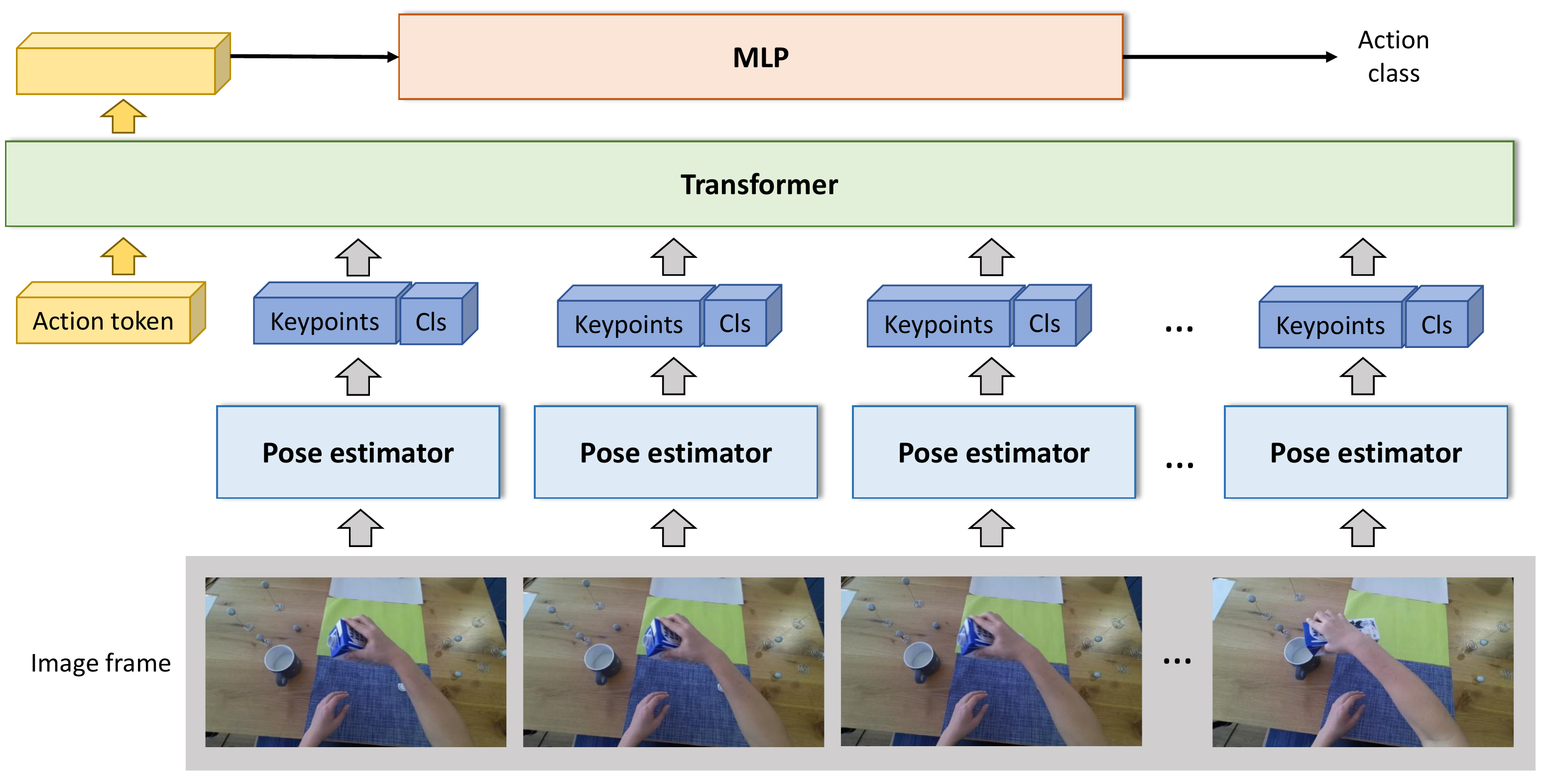}
\caption{The overall architecture of our Transformer-based model.}
\label{fig:model}
\end{figure*}

\section{Methods}
\label{sec:intro}
In this section, we introduce our proposed action recognition pipleine in Figure~\ref{fig:model} and explain details for solving problems.

\subsection{Keypoint Estimation based on Transformer}
We used DETR~\cite{carion2020end} as the baseline architecture to predict the 3D keypoints of two hands and an object. The object keypoints were defined as 8 corners, 12 edge midpoints and the centroid of the 3D bounding box. DETR is an architecture based on the Transformer's encoder-decoder structure, which performs object detection well without using pose-processing such as anchor generation or non-maximal suppression (NMS). We modified the head of the DETR decoder to predict keypoints and classes for each query. Classes are defined based on the hand types (ie. left, right hand) and the object classes. We used bipartite matching in the same way as DETR~\cite{carion2020end}. Hungarian loss is also same with that of DETR~\cite{carion2020end} except that their L1 loss is extended to object keypoints (ie. 8 corners, 12 edge midpoints and centroid of 3D bounding box) for object case and is extended to hand keypoint coordinates for left and right hand cases, respectively.

\subsection{Action Recognition}
As an action classifier, we involved another Transformer architecture which inherently offers the ability to deal with the sequential data. We followed the Vision Transformer~\cite{arnab2021vivit,dosovitskiy2020image} structure for this. First, the keypoints estimator predicts keypoints and classes. For keypoints of the left hand, right hand, and the object, the query with the highest probability of each class was employed for further stage. The learnable action token was concatenated with the results of each frame and putted as the input of the Transformer. The action token is updated through the self-attention mechanism, and the action token of the last layer predicts the action class through the multi-layer perceptron (MLP). Regarding the video sampling, we divided a video into $N$ clips and randomly selected frames from each clip for training. We used the cross-entropy loss to train the network for the action classes.

\section{Experiments}
\noindent \textbf{Implementation Details.}
The input image size is $960 \times 540$. The random horizontal flip was applied as a data augmentation. The number of clips per video is set as $64$ and the number of Transformer layer is set as $3$. We used the AdamW optimizer with learning rate of $10^{-4}$, and set the weight decay as $10^{-4}$. We used 1 RTX 3090 GPU and the batch size was set as $1$. Training takes $50$ epochs with a learning rate drop by a factor of $10$ after $40$ epochs.

\noindent \textbf{Evaluation Metrics.}
The results are evaluated with the top-1 accuracy metric. The top-1 accuracy is the conventional accuracy, model prediction must be exactly the expected ground-truth.

\begin{figure*}[!t]
\centering
\includegraphics[width=1\linewidth]{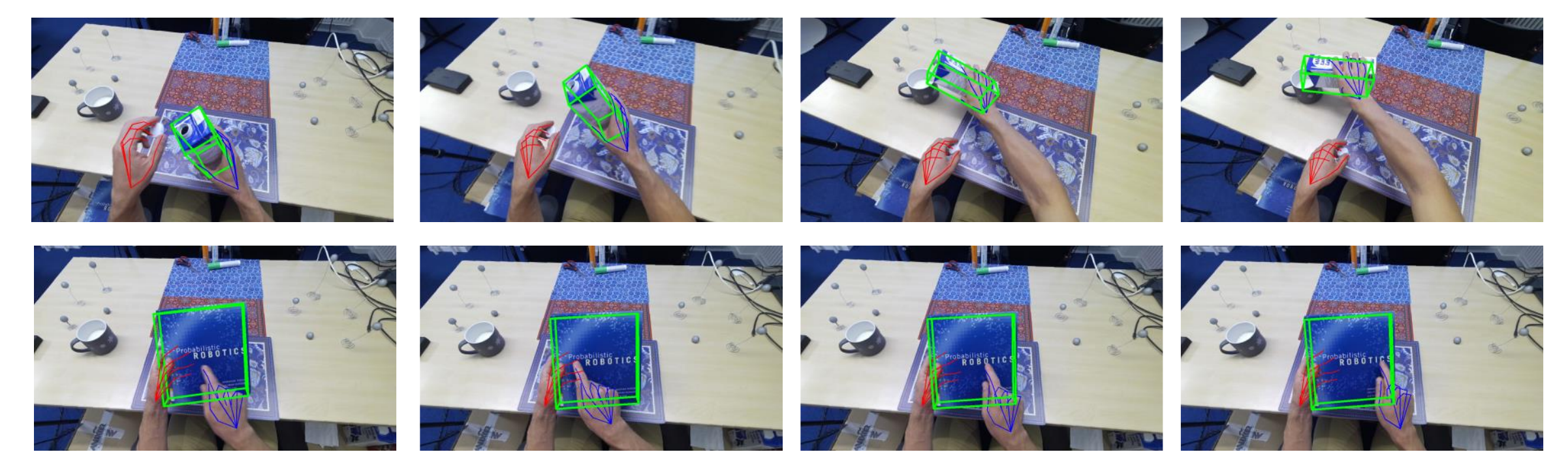}
\caption{Qualitative Results of our pose estimator on H2O dataset. The first row corresponds to pour milk, and the second row corresponds to a read book.}
\label{fig:vis}
\end{figure*}

\setlength{\tabcolsep}{4pt}
\begin{table}
\begin{center}

\label{table:headings}
\begin{tabular}{lll}
\hline\noalign{\smallskip}
method & Val accuracy(\%) & Test accuracy(\%)\\
\noalign{\smallskip}
\hline
\noalign{\smallskip}
C2D~\cite{wang2018non} & 76.10 & 70.66\\
I3D~\cite{carreira2017quo} & 85.15 & 75.21\\
SlowFast~\cite{feichtenhofer2019slowfast} & 86.00 & 77.69\\
H+O~\cite{tekin2019h+} & 80.49 & 68.88\\
H2O~\cite{kwon2021h2o} & 86.78 & 79.25\\
Ours & \textbf{90.87} & \textbf{87.19}\\

\hline
\end{tabular}
\end{center}
\caption{Comparison with state-of-the-art methods for H2O validation and test set}
\label{table:comparison}
\end{table}
\setlength{\tabcolsep}{1.4pt}

\setlength{\tabcolsep}{4pt}
\begin{table}
\begin{center}

\label{table:headings}
\begin{tabular}{llll}
\hline\noalign{\smallskip}
Sampling/train & Sampling/test & Frame &Test accuracy(\%)\\
\noalign{\smallskip}
\hline
\noalign{\smallskip}
uniform & uniform & 32 &83.40\\
uniform & uniform & 64 &84.71\\
N clips & N clips & 64 &86.36\\
N clips & uniform & 64 &\textbf{87.19}\\
\hline
\end{tabular}
\end{center}
\caption{Ablation study for H2O test set}
\label{table:comparison}
\end{table}
\setlength{\tabcolsep}{1.4pt}

\subsection{Evaluation Results}
As can be seen from Table~\ref{table:comparison}, our method outperforms the state-of-the-art method in both H2O validation set and test set. We conducted an ablation study on frame sampling during the train/test stages. There are two sampling methods: uniform sampling and segment random sampling. As can be seen in Table~\ref{table:comparison}, the best performance is obtained for the method that divides videos into clips, performs random sampling during training and performs uniform sampling during testing. Also, more frames the method uses, the better the accuracy becomes.

\section{Conclusion}
In this document, we described our 2nd place solution to ECCV 2022 challenge on Human Body, Hands, and Activities (HBHA) from Egocentric and Multi-view Cameras: Action Recognition. 

\bibliographystyle{splncs04}
\bibliography{main}
\end{document}